\documentclass[letterpaper]{article} 
\usepackage{aaai23}  
\usepackage{times}  
\usepackage{helvet}  
\usepackage{courier}  
\usepackage[hyphens]{url}  
\usepackage{graphicx} 
\usepackage{natbib}  
\usepackage{caption} 
\usepackage{algorithm}
\usepackage{algorithmic}

\usepackage{amsmath,amssymb} 
\usepackage{dsfont}
\usepackage{makecell}
\newcommand{\etal}{\textit{et al}.}
\usepackage{booktabs}
\usepackage{newfloat}
\usepackage{listings}
\DeclareCaptionStyle{ruled}{labelfont=normalfont,labelsep=colon,strut=off} 
\floatstyle{ruled}
\newfloat{listing}{tb}{lst}{}
\floatname{listing}{Listing}
\title{Automatic Neural Network Pruning that Efficiently Preserves the Model Accuracy}
\author{
    Thibault Castells and
    Seul-Ki Yeom\thanks{Corresponding author.}
}
\affiliations{
    Nota AI GmbH\\
    Friedrichstrasse 200, 10117 Berlin, Germany\\
    thibault@nota.ai, skyeom@nota.ai
}

\usepackage{bibentry}

\begin{document}

\maketitle

\begin{abstract}
    Neural networks performance has been significantly improved in the last few years, at the cost of an increasing number of floating point operations (FLOPs). When computational resources are limited, more FLOPs becomes an issue. As an attempt to solve this problem, pruning filters is a common solution, but most existing pruning methods do not preserve the model accuracy efficiently and therefore require a large number of finetuning epochs.
    In this paper, we propose an automatic pruning method that learns which neurons to preserve in order to maintain the model accuracy while reducing the FLOPs to a predefined target. To accomplish this task, we introduce a trainable bottleneck that only requires 25.6\% (CIFAR-10) or 15.0\% (ILSVRC2012) of the dataset within one single epoch to learn which filters to prune.
    Experiments on various architectures and datasets show that the proposed method can not only preserve the accuracy after pruning but also outperform existing methods after finetuning. With 52.00\% FLOPs reduction on ResNet-50, we achieve a Top-1 accuracy of 47.51\% after pruning and a state-of-the-art (SOTA) accuracy of 76.63\% after finetuning on ILSVRC2012.
    Code available at ~\url{https://github.com/nota-github/autobot_AAAI23}.
\end{abstract}

\section{Introduction}
\label{sec:intro}
In the last decade, Deep Neural Networks (DNNs) popularity has grown exponentially as the results improved, and they are now used in a variety of applications such as classification, detection, etc. However, these improvements are often faced with increasing model complexity, resulting in a need for more computational resources. Various attempts to make heavy models more compact have been proposed, based on different compression methods such as knowledge distillation~\cite{PolinoPA18,GuoWWYLHL20}, pruning~\cite{l1norm, hrank, ABCPruner, LRP_pruning}, quantization~\cite{QuZCT20}, neural architecture search (NAS)~\cite{hournas}, \textit{etc}.
Network pruning, which consists in removing redundant and unimportant connections, received great interest from the industry as it is a simple and effective solution. While the main challenge of this method is to find a good pruning criterion, another difficulty is to define what percentage of each layer should be pruned. As a manual search is a time-consuming process that requires human expertise, recent works have proposed methods that automatically prune the redundant filters throughout the network to meet a given constraint such as the number of parameters, FLOPs, or hardware platform~\cite{liunetworkslimming, you2019gate, Li_2021_CVPR, importance_estimation, ABCPruner, LRP_pruning, nisp, Info_flow18, Info_flow21_PAMI}. To automatically find the best-pruned architectures, these methods rely on various metrics such as the 2nd order Taylor expansions~\cite{importance_estimation}, the layer-wise relevance propagation score~\cite{LRP_pruning}, \textit{etc}. For further details, please find Sec.~\ref{sec:relatedwork}. Although these strategies improved over time, they usually do not explicitly aim to preserve the model accuracy, or they do it in a computationally expensive way.

\begin{figure}[t]
\centering
\includegraphics[width=1\columnwidth]{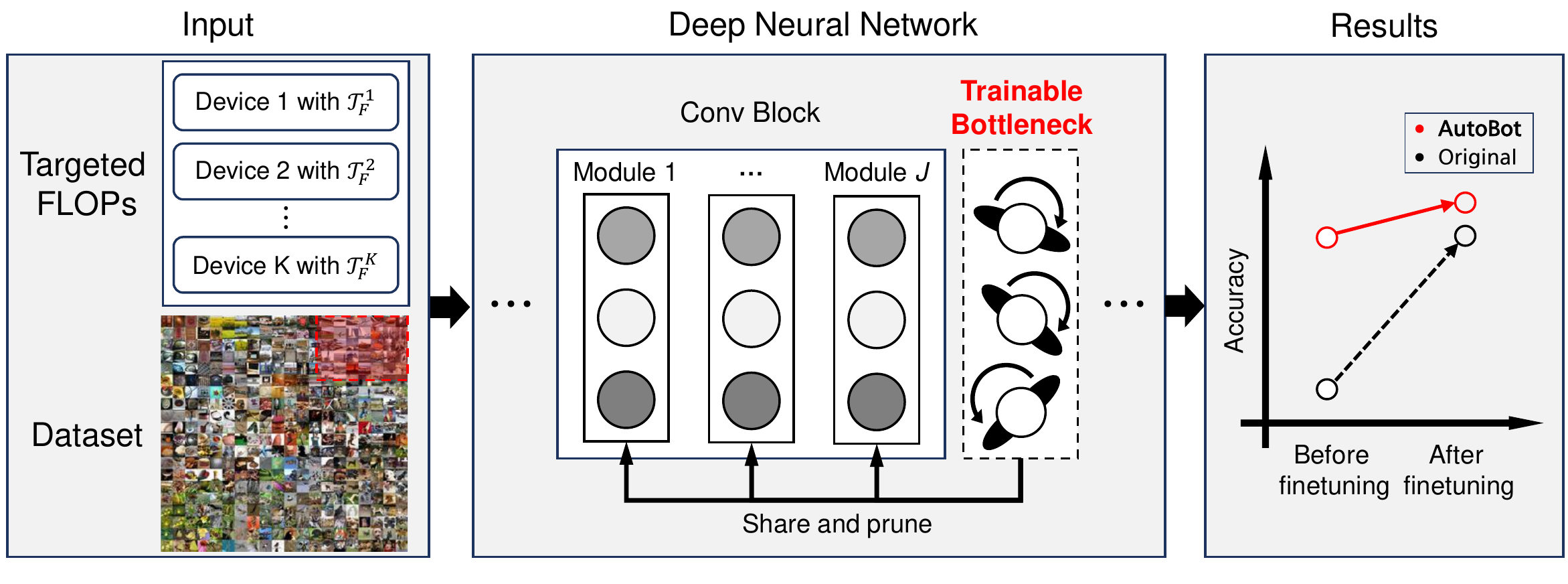}
\caption{System flow of AutoBot for automatic network pruning. A trainable bottleneck is injected after each convolution block. They are then updated to restrict the information flow (like water taps) while minimizing the accuracy drop, with the given targeted FLOPs and a small amount of data. The most restricted filters are pruned afterwards. As a result, compared to other existing pruning methods, AutoBot can efficiently preserve the accuracy, leading to a SOTA accuracy after finetuning.}
\label{fig:fig1}
\end{figure}

In this paper, we make the hypothesis that, for the same compression, the pruned architecture that can lead to the best accuracy after finetuning is the one that most efficiently preserves the accuracy during the pruning process (see Sec.~\ref{subsec:initial_acc}). We therefore introduce an automatic pruning method, called AutoBot, that uses trainable bottlenecks to efficiently preserve the model accuracy while minimizing the FLOPs, as shown in Fig.~\ref{fig:fig1}.
These bottlenecks only require one single epoch of training with 25.6\% (CIFAR-10) or 15.0\% (ILSVRC2012) of the dataset to efficiently learn which filters to prune. We compare AutoBot with various pruning methods, and show a significant improvement of the pruned models before finetuning, leading to a SOTA accuracy after finetuning. We also perform a practical deployment test on several edge devices to demonstrate the speed improvement of the pruned models.

To summarize, our contributions are as follows:\baselinestretch
\begin{itemize}
    \item We introduce AutoBot, a novel automatic pruning method that uses a trainable bottleneck to efficiently learn which filter to prune in order to maximize the accuracy while minimizing the FLOPs of the model. This method can easily and intuitively be implemented regardless of the dataset or model architecture.
    \item We demonstrate that preserving the accuracy during the pruning process has a strong impact on the accuracy of the finetuned model (Sec.~\ref{subsec:initial_acc}).
    \item Extensive experiments show that AutoBot efficiently preserve the accuracy after pruning (before finetuning), and outperforms previous pruning methods once finetuned.
\end{itemize}



\section{Related Works}
\label{sec:relatedwork}
In this section, we summarize some related works compared to our proposed method.
Traditionally, magnitude-based pruning aims to exploit the inherent characteristics of the network to define a pruning criterion, without modifying the network parameters. Popular criteria include $l_p$-norm~\cite{l1norm, Minbao2021DCFF, Li2020CVPR}, Taylor expansion~\cite{importance_estimation}, Gradient~\cite{LiuWgradient19}, Singular Value Decomposition~\cite{hrank}, sparsity of output feature maps~\cite{apoz}, geometric median~\cite{GM}, etc. Recently,~\citet{SCOP} proposed a scientific control pruning method, called SCOP, which introduces knockoff features as the control group. In contrast, adaptive pruning needs to retrain the networks from scratch with a modified training loss or architecture which adds new constraints. Several works~\cite{liunetworkslimming, luo2017thinet, YeRethinking18} add trainable parameters to each feature map channel to obtain data-driven channel sparsity, enabling the model to automatically identify redundant filters. \citet{luo2017thinet} introduce Thinet that formally establishes filter pruning as an optimization problem and prunes filters based on statistical information computed from its next layer, not the current layer. \citet{GAL} propose a structured pruning method that jointly prunes filters and other structures by introducing a soft mask with sparsity regularization. However, retraining the model from scratch is a time- and resource-consuming process that does not significantly improve the accuracy compared to magnitude-based pruning. Although these two pruning strategies are intuitive, the pruning ratio must be manually defined layer-by-layer, which is a time-consuming process that requires human expertise. Instead, in this paper, we focus on automatic pruning.

As suggested by the name, automatic network pruning removes the redundant filters throughout the network automatically under any constraints such as a number of parameters, FLOPs, or hardware platform. In this respect, a large number of automatic pruning methods have been proposed.
\citet{liunetworkslimming} optimize the scaling factor $\gamma$ in the batch-norm layer as a channel selection indicator to decide which channels are unimportant. 
\citet{you2019gate} propose an automatic pruning method, called Gate Decorator, which transforms CNN modules by multiplying their output by channel-wise scaling factors and adopt an iterative pruning framework called Tick-Tock to boost pruning accuracy. \citet{Li_2021_CVPR} propose a collaborative compression method that mutually combines channel pruning and tensor decomposition.
\citet{importance_estimation} estimates the contribution of a filter to the final loss using 2nd order Taylor expansions and iteratively removes those with smaller scores.
\citet{ABCPruner} propose ABCPruner to find the optimal pruned structure automatically by updating the structure set and recalculating the fitness.
Back-propagation methods~\cite{LRP_pruning, nisp} compute the relevance score of each filter by following the information flow from the model output. 
\citet{Info_flow18} and~\citet{Info_flow21_PAMI} adopt information theory to preserve the information between the hidden representation and input or output.

Most existing methods are computationally and time expensive because they either require to retrain the model from scratch~\cite{liunetworkslimming}, apply iterative pruning~\cite{you2019gate, importance_estimation, LRP_pruning, nisp, Li_2021_CVPR} or finetune the model while pruning~\cite{ABCPruner, Info_flow18}. When the model isn't retrained or finetuned during the pruning process, they generally do not preserve the model accuracy after pruning~\cite{Info_flow21_PAMI, LRP_pruning, nisp}, and thus require to be finetuned for a large number of epochs.
In contrast to other automatic pruning methods, AutoBot stands out by its speed and its ability to preserve the accuracy of the model during the pruning process.

\begin{algorithm}[tb]
    \caption{$AutoBot$}
    \label{alg:algo1}
    \textbf{Input}: pre-trained model $f$, targeted FLOPs $F_{T}$, acceptable FLOPs error $\epsilon$, hyper-parameter $\beta$, number of iterations $k$, training data $D$\\
    \textbf{Output}: Pruned model $f'$
    \begin{algorithmic}[1] 
        \STATE Inject $Trainable$ $Bottlenecks$ in $f$
        \FOR{Batch $\mathcal{X}$ in $D[0; k]$} 
            \STATE $\mathcal{L}$ $\gets$ $\mathcal{L}_{CE}(f(\mathcal{X};\mathbf{\Lambda})) + \beta \mathcal{L}_{g}$($\mathbf{\Lambda}$)
            \STATE $\mathbf{\Lambda}$ $\gets$ $Update$($\mathbf{\Lambda}$, $\mathcal{L}$)
        \ENDFOR
        \STATE $\mathbf{\Lambda}_{bool}$ $\gets$ $GetPruningMask$($\mathbf{\Lambda}$, $F_{T}$, $\epsilon$)
        \STATE Remove $Trainable$ $Bottlenecks$ from $f$
        \STATE $f'$ $\gets$ $Prune$($f, \mathbf{\Lambda}_{bool}$) 
        \STATE $f'$ $\gets$ $Finetune$($f'$, $D$)
        \STATE \textbf{return} $f'$
    \end{algorithmic}
\end{algorithm}

\begin{algorithm}[tb]
    \caption{$GetPruningMask$}
    \label{alg:algo2}
    \textbf{Input}: trained bottlenecks parameters $\mathbf{\Lambda}$, targeted FLOPs $F_{T}$, acceptable FLOPs error $\epsilon$\\
    \textbf{Output}: pruning mask $\mathbf{\Lambda}_{bool}$
    \begin{algorithmic}[1] 
        \STATE $\mathcal{T}$ $\gets$ $0.5$
        \STATE $\mathbf{\Lambda}_{bool}$ $\gets$ $1$ where $\mathbf{\Lambda} > \mathcal{T}$, $0$ elsewhere
        \STATE $F$ $\gets$ g$(\mathbf{\Lambda}_{bool})$ \qquad (Eq.~\ref{eq:FLOPs_func})
        \STATE $i$ $\gets$ $0$
        \WHILE{$\left| F - F_{T} \right| > \epsilon$}
            \IF{$F > F_{T}$}
                \STATE $\mathcal{T}$ $\gets$ $\mathcal{T} + \frac{0.25}{2^i}$ 
            \ELSE
                \STATE $\mathcal{T}$ $\gets$ $\mathcal{T} - \frac{0.25}{2^i}$ 
            \ENDIF
            \STATE $\mathbf{\Lambda}_{bool}$ $\gets$ $1$ where $\mathbf{\Lambda} > \mathcal{T}$, $0$ elsewhere
            \STATE $F$ $\gets$ g$(\mathbf{\Lambda}_{bool})$
            \STATE $i$ $\gets$ $i+1$ 
        \ENDWHILE
        \STATE \textbf{return}  $\mathbf{\Lambda}_{bool}$
    \end{algorithmic}
\end{algorithm}

\section{Method}
\label{sec:method}
Motivated by several bottleneck approaches~\cite{info_bottleneck,deep_info_bottleneck,schulz2020iba}, our method control the information flow throughout the pretrained network using \textit{Trainable Bottlenecks} that are injected into the model. 
The objective function of the trainable bottleneck is to maximize the information flow from input to output while minimizing the loss by adjusting the amount of information in the model under the given constraints. During the training procedure, only the parameters $\mathbf{\Lambda}$ of the trainable bottlenecks are updated while all the pretrained parameters of the model are frozen.

Compared to other pruning methods inspired by the information bottleneck~\cite{Info_flow18, Info_flow21_PAMI}, we do not consider the compression of mutual information between the input/output and the hidden representations in order to evaluate the information flow. Such methods are orthogonal to AutoBot, which explicitly quantifies how much information is passed through each layer. This explicit quantification result in a faster training --we optimize the trainable bottlenecks on a fraction of one single epoch only-- and an improved capacity to preserve the accuracy. Our AutoBot pruning process is summarized in Alg.~\ref{alg:algo1}.

\subsection{Trainable Bottleneck}
We formally define the concept of trainable bottleneck as an operator that can restrict the information flow throughout the network during the forward pass, using trainable parameters. Mathematically, it can be formulated as:
\begin{equation}
    \label{eq:bottleneck_general}
    X_{i+1} = B(\boldsymbol{\lambda}_i, X_{i})\\
\end{equation}
\noindent where $B$ stands for the trainable bottleneck, $\boldsymbol{\lambda}_i$ denotes the bottleneck parameters of the $i^{th}$ operator, and $X_{i}$ and $X_{i+1}$ denote the input and output feature map of the bottleneck at the $i^{th}$ operator, respectively. For instance, Schulz \etal~\cite{schulz2020iba} control the amount of information into the model by injecting noise into it. In this case, $B$ is expressed as $B(\boldsymbol{\lambda}_i, X_{i}) = \boldsymbol{\lambda}_i X_{i} + (1 - \boldsymbol{\lambda}_i) \epsilon$ where $\epsilon$ denotes the noise. 

Inspired by the information bottleneck concept~\cite{info_bottleneck, deep_info_bottleneck}, we formulate a general bottleneck that is not limited to only information theory but can be optimized to satisfy any constraint as follow:
\begin{align}
    \label{eq:objective_function}
    \begin{split}
    \min_{\mathbf{\Lambda}} \mathcal{L}_{CE} (\mathcal{Y}, f(\mathcal{X};\mathbf{\Lambda})) \qquad & s.t. \quad r(\mathbf{\Lambda}) \leq \mathcal{C}
    \end{split}
\end{align}
\noindent where $\mathcal{L}_{CE}$ stands for the cross-entropy loss, $\mathcal{X}$ and $\mathcal{Y}$ stand for the model input and output, $\mathbf{\Lambda}$ is the set of the bottleneck parameters  ($\mathbf{\Lambda} = [ \boldsymbol{\lambda}_1, \boldsymbol{\lambda}_2, \ldots, \boldsymbol{\lambda}_L ]$) in the model, $r$ is a constraint function, and $\mathcal{C}$ is the desired constraint.

\begin{figure}[t]
\centering
\includegraphics[width=0.75\columnwidth]{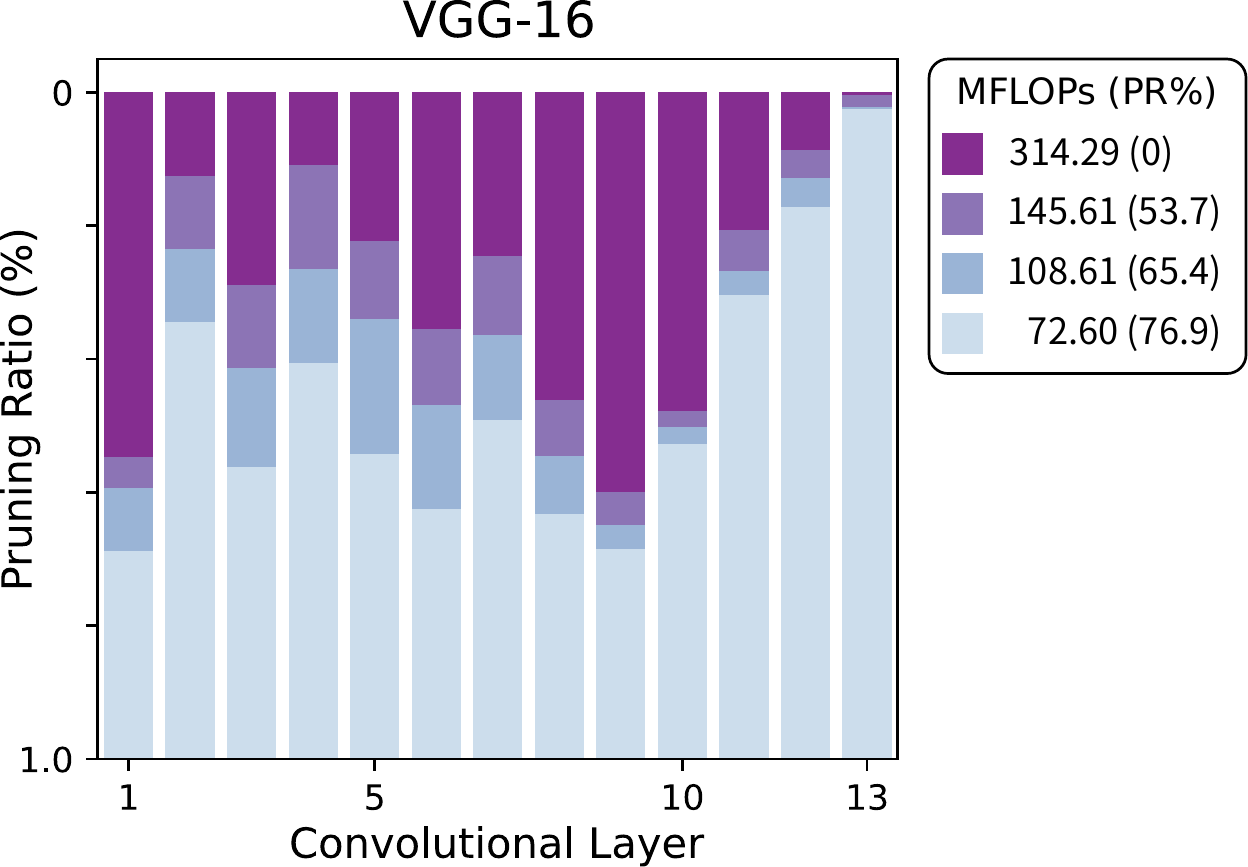}
\caption{Per-layer filter pruning ratio for various targeted FLOPs on VGG-16. This ratio is automatically determined by AutoBot to satisfy the targeted FLOPs.}
\label{fig:fig2}
\end{figure}

\subsection{Pruning Strategy}
\label{subsec:pruning_strategy}

In the following, we define a convolution block as a convolution layer, plus all the following operators that preserve the number and order of channels. It can contain multiple convolutions if their outputs are merged (in the case of a skip connection).
In this work, we inject a bottleneck into each convolution block throughout the network such that the information flow of the estimated model to be pruned is quantified by restricting trainable parameters layer-wisely.\\

Compared to previous works, our bottleneck function $B(\boldsymbol{\lambda}_i, X_{i})$ (Eq.~\ref{eq:bottleneck_general}) do not use noise to control the information flow:
\begin{equation}
    \label{eq:bottleneck}
    X_{i+1} = \boldsymbol{\lambda}_i X_{i}\\
\end{equation}
where $\boldsymbol{\lambda}_i \in [0,1]$. Therefore the range of $X_{i+1}$ is changing from $[\epsilon, X_{i}]$ to $[0, X_{i}]$. For pruning, this is more relevant since replacing an operator input by zeros is equivalent to pruning the operator (i.e. pruning the corresponding output of the previous operator).\\

Following the general objective function of the trainable bottleneck (Eq.~\ref{eq:objective_function}), we introduce a regularizer $g$ to constrain the FLOPs of the pruned architecture:
\begin{align}
    \label{eq:objective_function_autobot}
    \begin{split}
    \min_{\mathbf{\Lambda}} \mathcal{L}_{CE} (\mathcal{Y}, f(\mathcal{X};\mathbf{\Lambda})) \qquad & s.t. \quad g(\mathbf{\Lambda}) = \mathcal{T}_{F}
    \end{split}
\end{align}
where $\mathcal{T}_{F}$ is the target FLOPs (manually fixed), and $g(\mathbf{\Lambda})$ estimates the FLOPs of the model weighted by $\Lambda$. Formally, given a neural network consisting of multiple convolutional blocks, we define $g$ as follows:
\begin{equation}
    \label{eq:FLOPs_func}
    g(\mathbf{\Lambda}) = \sum_{i=1}^{L} \sum_{j=1}^{J_{i}} {g_{i}^{j}(\boldsymbol{\lambda}_{i}, \boldsymbol{\lambda}_{i-1})}
\end{equation}
where $\boldsymbol{\lambda}_{i}$ is the vector of parameters of the information bottleneck following the $i^{th}$ convolution block, $g_{i}^{j}$ is the function that computes the FLOPs of the $j^{th}$ operator of the $i^{th}$ convolution block weighted by $\boldsymbol{\lambda}_{i}$, $L$ is the total number of convolution blocks in the model and $J_{i}$ is the total number of operators in the $i^{th}$ convolution block. 
For instance, if $g_{i}^{j}$ is for a convolutional operator without bias and padding, it is expressed as:
\begin{equation}
    \label{eq:FLOPs_func_example}
    g_{i}^{j}(\boldsymbol{\lambda}_{i}, \boldsymbol{\lambda}_{i-1}) = sum(\boldsymbol{\lambda}_{i}) \times sum(\boldsymbol{\lambda}_{i-1}) \times h \times w \times k \times k
\end{equation}
where $h$ and $w$ are the height and width of the output feature map of the convolution, and $k$ is its kernel size.
Notice that within the $i^{th}$ convolution block, all operators share $\boldsymbol{\lambda}_{i}$. That is, at a block level all the operators belonging to the same convolution block are pruned together.

\begin{table*}[!ht]
    \centering
    \caption{Pruning results of five network architectures on CIFAR-10, sorted by FLOPs in descending order. Scores in brackets denote the pruning ratio in the compressed models. Unless specified otherwise, the accuracy before finetuning was re-computed by us using the code from the corresponding paper.}
    \label{tab:cifar10_table}
    \scalebox{0.73}{
        \begin{tabular}{ccccccc}
            \specialrule{1pt}{1pt}{1pt}
            \hline
            Method          & Automatic & Top1-acc           &  Top1-acc & $\uparrow$~$\downarrow$ &        FLOPs         &     Params      \\ 
                            &           & before finetuning  &           &                         &    (Pruning Ratio)   & (Pruning Ratio) \\ \hline
            VGG-16~\cite{vgg}                  &  &       --    & 93.96\%            & 0.0\%             & 314.29M (0.0\%)              & 14.99M (0.0\%)                     \\
            L1~\cite{l1norm}                       & &       88.70\%\textsuperscript{**}    & 93.40\%  & -0.56\%           & 206.00M (34.5\%)             & 5.40M (64.0\%)    \\
            CC-0.5~\cite{Li_2021_CVPR}        & \checkmark & --    & 94.15\%            & +0.19\%           & 154.00M (51.0\%)             & 5.02M (66.5\%)                  \\
            \textbf{AutoBot (Ours)}                  & \checkmark & \textbf{88.29\%}    & \textbf{94.19\%}   & \textbf{+0.23\%}  & \textbf{145.61M (53.7\%)}    & \textbf{7.53M (49.8\%)}      \\
            CC-0.6~\cite{Li_2021_CVPR}        & \checkmark & --    & 94.09\%            & +0.13\%           & 123.00M (60.9\%)             & 5.02M (73.2\%)                             \\
            HRank-65~\cite{hrank}                  & &       10.06\%    & 92.34\%            & -1.62\%           & 108.61M (65.4\%)             & 2.64M (82.4\%)    \\
            \textbf{AutoBot (Ours)}                  & \checkmark & \textbf{82.73\%}    & \textbf{94.01\%}   & \textbf{+0.05\%}  & \textbf{108.71M (65.4\%)}    & \textbf{6.44M (57.0\%)}      \\
            ITPruner~\cite{Info_flow21_PAMI}       & \checkmark &       10.00\%\textsuperscript{*}    & 94.00\%            & +0.04\%           & 98.80 (68.6\%)               & --      \\
            ABCPruner~\cite{ABCPruner} & \checkmark &   10.00\%\textsuperscript{*}    & 93.08\%            & -0.88\%           & 82.81M (73.7\%)              & 1.67M (88.9\%)     \\
            DCFF~\cite{Minbao2021DCFF}             & &       --    & 93.49\%            & -0.47\%           & 72.77M (76.8\%)              & 1.06M (92.9\%)     \\
            \textbf{AutoBot (Ours)}                & \checkmark & \textbf{71.24\%}      & \textbf{93.62\%}   & \textbf{-0.34\%}  & \textbf{72.60M (76.9\%)}     & \textbf{5.51M (63.24\%)}  \\
            VIBNet~\cite{Info_flow18}       & \checkmark &   --    & 91.50\%            & -2.46\%           & 70.63M (77.5\%)              & -- (94.7\%)         \\ \hline

            ResNet-56~\cite{resNet}                & &       --    & 93.27\%            & 0.0\%             & 126.55M (0.0\%)            & 0.85M (0.0\%)       \\
            L1~\cite{l1norm}                       & &       --    & 93.06\%            & -0.21\%           & 90.90M (28.2\%)            & 0.73M (14.1\%)      \\
            HRank-50~\cite{hrank}                  & & 10.73\% & 93.17\%  & -0.10\%     & 62.72M (50.4\%)            & 0.49M (42.4\%)      \\
            SCP~\cite{KangH20}                     & &       --    & 93.23\%            & -0.04\%           & 61.89M (51.1\%)            & 0.44M (48.2\%)      \\
            CC~\cite{Li_2021_CVPR}           & \checkmark &  26.54\%     & 93.64\%            & +0.37\%           & 60.00M (52.6\%)            & 0.44M (48.2\%)      \\
            ITPruner~\cite{Info_flow21_PAMI}       & \checkmark &    10.00\%\textsuperscript{*}    & 93.43\%            & +0.16\%           & 59.50 (53.0\%)             & --       \\
            FPGM~\cite{GM}                         & &       17.44\%    & 93.26\%            & -0.01\%           & 59.40M (53.0\%)            & --                  \\
            LFPC~\cite{Cpruning_variousCriteria}   & &       --    & 93.24\%            & -0.03\%           & 59.10M (53.3\%)            & --                  \\
            ABCPruner~\cite{ABCPruner} & \checkmark &   10.00\%\textsuperscript{*}    & 93.23\%            & -0.04\%           & 58.54M (53.7\%)              & 0.39M (54.1\%)    \\
            DCFF~\cite{Minbao2021DCFF}             & &       --    & 93.26\%            & -0.01\%           & 55.84M (55.9\%)            & 0.38M (55.3\%)      \\
            \textbf{AutoBot (Ours)}             & \checkmark & \textbf{85.58\%}    & \textbf{93.76\%}         & \textbf{+0.49\%}       & \textbf{55.82M (55.9\%)}   & \textbf{0.46M (45.9\%)}  \\ 
            SCOP~\cite{SCOP}                       & &       57.34\%    & 93.64\%            & +0.37\%           & -- (56.0\%)                & -- (56.3\%)         \\  \hline

            ResNet-110~\cite{resNet}               & &       --    & 93.5\%             & 0.0\%             & 254.98M (0.0\%)             & 1.73M (0.0\%)       \\
            L1~\cite{l1norm}                       & &       --    & 93.30\%            & -0.20\%           & 155.00M (39.2\%)              & 1.16M (32.9\%)    \\
            FPGM~\cite{GM}                       & &       11.79\%    & 93.74\%            & +0.24\%           & 121.00M (52.5\%)              & --    \\
            HRank-58~\cite{hrank}                  & &       10.00\%    & 93.36\%            & -0.14\%           & 105.70M (58.5\%)              & 0.70M (59.5\%)    \\
            LFPC~\cite{Cpruning_variousCriteria}   & &       --    & 93.07\%            & -0.43\%           & 101.00M (60.3\%)              & --                \\
            ABCPruner~\cite{ABCPruner} & \checkmark &   10.00\%\textsuperscript{*}    & 93.58\%            & +0.08\%           & 89.87M (64.8\%)               & 0.56M (67.6\%)    \\
            DCFF~\cite{Minbao2021DCFF}             & &       --    & 93.80\%            & +0.30\%           & 85.30M (66.5\%)               & 0.56M (67.6\%)    \\
            \textbf{AutoBot (Ours)}                 & \checkmark & \textbf{84.37\%}    & \textbf{94.15\%}     & \textbf{+0.65\%}      & \textbf{85.28M (66.6\%)}      & \textbf{0.70M (59.5\%)}                 \\ \hline

            GoogLeNet~\cite{googlenet}             & &       --    & 95.05\%         & 0.0\%             & 1.53B (0.0\%)               & 6.17M (0.0\%)                \\ 
            L1~\cite{l1norm}                       & &       --    & 94.54\%           & -0.51\%           & 1.02B (33.3\%)              & 3.51M (43.1\%)                 \\
            Random                                 & &       10.00\%    & 94.54\%           & -0.51\%           & 0.96B (37.3\%)              & 3.58M (42.0\%)                 \\
            HRank-54~\cite{hrank}                  & &       10.00\%    & 94.53\%           & -0.52\%           & 0.69B (54.9\%)              & 2.74M (55.6\%)                 \\
            CC~\cite{Li_2021_CVPR}          & \checkmark &   --    & 94.88\%           & -0.17\%           & 0.61M (60.1\%)             & 2.26M (63.4\%)                  \\
            ABCPruner~\cite{ABCPruner} & \checkmark &   10.00\%\textsuperscript{*}    & 94.84\%           & -0.21\%           & 0.51B (66.7\%)              & 2.46M (60.1\%)                 \\
            DCFF~\cite{Minbao2021DCFF}             & &       --    & 94.92\%           & -0.13\%           & 0.46B (69.9\%)              & 2.08M (66.3\%)                 \\
            HRank-70~\cite{hrank}                  & &       10.00\%    & 94.07\%           & -0.98\%           & 0.45B (70.6\%)              & 1.86M (69.9\%)                 \\
            \textbf{AutoBot (Ours)}                 & \checkmark & \textbf{90.18\%}    & \textbf{95.23\%}   & \textbf{+0.16\%}   & \textbf{0.45B (70.6\%)}     & \textbf{1.66M (73.1\%)}       \\ \hline

            DenseNet-40~\cite{densenet}            & &       --    & 94.81\%        & 0.0\%             & 287.71M (0.0\%)               & 1.06M (0.0\%)                \\
            GAL-0.01~\cite{GAL}                    & &       --    & 94.29\%        & -0.52\%           & 182.92M (36.4\%)              & 0.67M (36.8\%)                 \\
            \textbf{AutoBot (Ours)}                       & \checkmark & \textbf{87.85\%} & \textbf{94.67\%} & \textbf{-0.14\%} & \textbf{167.64M (41.7\%)}  & \textbf{0.76M (28.3\%)} \\
            HRank-40~\cite{hrank}    & & 25.58\%  & 94.24\%        & -0.57\%           & 167.41M (41.8\%)              & 0.66M (37.7\%)                 \\
            Variational CNN~\cite{zhao2019variational}& &       --    & 93.16\%        & -1.65\%           & 156.00M (45.8\%)              & 0.42M (60.4\%)                 \\
            \textbf{AutoBot (Ours)}                       & \checkmark & \textbf{83.20\%}    & \textbf{94.41\%}    & \textbf{-0.4\%}   & \textbf{128.25M (55.4\%)}  & \textbf{0.62M (41.5\%)}   \\ 
            GAL-0.05~\cite{GAL}                    & &       --    & 93.53\%        & -1.28\%           & 128.11M (55.5\%)              & 0.45M (57.5\%)                 \\
            \specialrule{1pt}{1pt}{1pt} 
            \midrule[.5pt]
            \multicolumn{6}{l}{\textsuperscript{*}\footnotesize{this method train the pruned model from scratch, instead of finetuning}} \\
            \multicolumn{6}{l}{\textsuperscript{**}\footnotesize{according to~\cite{neuron_merging}}} 
        \end{tabular}
    }
\end{table*}

To solve our optimization problem defined in Eq.~\ref{eq:objective_function_autobot}, we introduce $\mathcal{L}_{g}$, a loss term designed to satisfy the constraint $g$ from Eq.~\ref{eq:FLOPs_func}. We formulate $\mathcal{L}_{g}$ as follow:

\begin{equation}
    \label{eq:flops_loss}
    \mathcal{L}_{g} =
    \begin{cases}
        \frac{g(\mathbf{\Lambda}) - \mathcal{T}_{F}}{\mathcal{M}_{F}-\mathcal{T}_{F}},& \text{if } g(\mathbf{\Lambda}) \geq \mathcal{T}_{F}\\
        \\
        1 - \frac{g(\mathbf{\Lambda})}{\mathcal{T}_{F}}, & \text{otherwise}
    \end{cases}
\end{equation}
where $\mathcal{M}_{F}$ is the FLOPs of the original model, and $\mathcal{T}_{F}$ is the predefined target FLOPs. 

In contrast to $g$, this loss term is normalized such that the scale of the loss is always the same. As a result, for a given dataset, the training parameters are stable across different architectures. The optimization problem to update the proposed information bottlenecks for automatic pruning can be summarized as follows:

\begin{equation}
    \label{eq:objective_function_final}
    \min_{\mathbf{\Lambda}} \mathcal{L}_{CE} (\mathcal{Y}, f(\mathcal{X};\mathbf{\Lambda})) + \beta \mathcal{L}_{g}(\mathbf{\Lambda})\\
\end{equation}
where $\beta$ is a hyper-parameter that indicates the relative importance of its associated objective.\\

\noindent
{\bf From $\mathbf{\Lambda}$ to pruning mask} Once the bottlenecks are trained, $\mathbf{\Lambda}$ can be directly used as a pruning criterion. Therefore, we propose a way to quickly find the threshold under which neurons should be pruned. Since our bottleneck allows us to quickly and accurately compute the weighted FLOPs (Eq.~\ref{eq:FLOPs_func}), we can estimate the FLOPs of the model to be pruned without actual pruning. This is done by setting $\mathbf{\Lambda}$ to zero for the filters to be pruned, or one otherwise. We call this process \textit{pseudo-pruning}.
In order to find the optimal threshold, we initialize a threshold to 0.5 and \textit{pseudo-prune} all filters with $\mathbf{\Lambda}$ lower than this threshold. We then compute the weighted FLOPs, and adopt the binary search algorithm to efficiently minimize the distance between the current and targeted FLOPs. This process is repeated until the gap is small enough. This process is summarized in Alg.~\ref{alg:algo2}. Once we have found the optimal threshold, we cut out all bottlenecks from the model and finally prune all the filters with $\mathbf{\Lambda}$ lower than this threshold to get the compressed model with the targeted FLOPs. This whole process takes less than a second on CPU as it is based on the binary search algorithm, which has a complexity of $\mathcal{O}(\log{}n)$, $n$ being the number of FLOPs in this case.\\

\noindent
{\bf Parametrization} Following~\citet{schulz2020iba}, we do not directly optimize $\mathbf{\Lambda}$ as this would require to use clipping to stay in the $[0, 1]$ interval. Instead, we parametrize $\mathbf{\Lambda} = \text{sigmoid}(\mathbf{\Psi})$, where the elements of $\mathbf{\Psi}$ are in $\mathbb{R}$.\\

\begin{figure}[t]
    \centering
    \includegraphics[width=0.95\columnwidth]{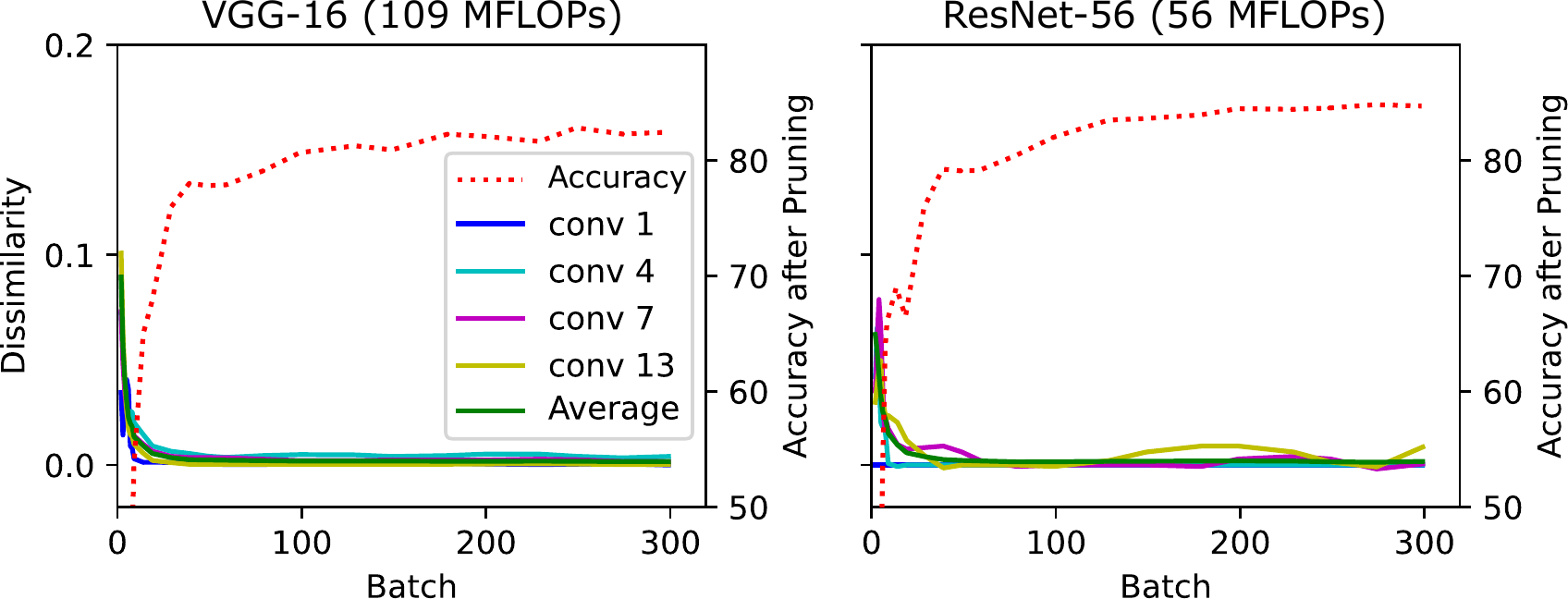}
    \caption{Evolution of accuracy after pruning (before finetuning) and of dissimilarity between filters ranking (normalised Kendall tau distance) when increasing the number of batches, on CIFAR-10.}
    \label{fig:acc_after_pruning_and_dissimilarity}
\end{figure}

\noindent
{\bf Reduced training data} We empirically observed that the training for the bottlenecks can converge quickly before the end of the first epoch. For instance, we can observe on Fig.~\ref{fig:acc_after_pruning_and_dissimilarity} that around 200 batches are needed (25.6\% of the dataset) to converge on CIFAR-10. For ILSVRC2012, the same observation is made with 15.0\% of the dataset. Therefore, it suggests that regardless of model size (i.e. FLOPs), the optimally pruned architecture can be efficiently estimated using only a small portion of the dataset.

\section{Experiments}
\label{sec:experiments}
\begin{table*}[ht]
    \caption{Pruning results on ResNet-50 with ImageNet, sorted by FLOPs. Scores in brackets of ``FLOPs'' and ``Params'' denote the pruning ratio of FLOPs and number of parameters in the compressed models. Accuracy before finetuning was re-computed by us using the code from the corresponding paper.}
    \centering
    \scalebox{0.73}{
        \begin{tabular}{cccccccc}
            \specialrule{1pt}{1pt}{1pt}
            \hline
            Method          & Automatic & Top1-acc           &  Top1-acc & $\uparrow$~$\downarrow$ & Top5-acc &        FLOPs         &     Params      \\ 
                            &           & before finetuning &           &                         &        &    (Pruning Ratio)   & (Pruning Ratio) \\ \hline

            ResNet-50~\cite{resNet}                 & & --             & 76.13\%             & 0.0\%             & 92.87\%           & 4.11B (0.0\%)      & 25.56M (0.0\%) \\
            ThiNet-50~\cite{luo2017thinet}          & & --             & 72.04\%             & -4.09\%           & 90.67\%           & -- (36.8\%)        & -- (33.72\%) \\
            FPGM~\cite{GM}                          & & 0.25\%             & 75.59\%             & -0.59\%           & 92.27\%       & 2.55B (37.5\%)     & 14.74M (42.3\%) \\
            ABCPruner~\cite{ABCPruner}    & \checkmark & 0.10\%\textsuperscript{*}      & 74.84\%       & -1.29\%     & 92.31\%      & 2.45B (40.8\%)     & 16.92M (33.8\%) \\
            SFP~\cite{l2norm}                       & & --             & 74.61\%             & -1.52\%           & 92.06\%           & 2.38B (41.8\%)     & -- \\
            HRank-74~\cite{hrank}                   & & 0.09\%             & 74.98\%             & -1.15\%           & 92.33\%       & 2.30B (43.7\%)     & 16.15M (36.8\%) \\
            Taylor~\cite{importance_estimation}             & & --             & 74.50\%             & -1.63\%           & --        & -- (44.5\%)        & -- (44.9\%)\\
            DCFF~\cite{Minbao2021DCFF}              & & --             & 75.18\%             & -0.95\%           & 92.56\%           & 2.25B (45.3\%)     & 15.16M (40.7\%) \\
            ITPruner~\cite{Info_flow21_PAMI} & \checkmark & 0.10\%\textsuperscript{*}  & 75.75\%     & -0.38\%   & --                & 2.23B (45.7\%)     & -- \\
            AutoPruner~\cite{luo2020autopruner} & \checkmark & --      & 74.76\%             & -1.37\%           & 92.15\%           & 2.09B (48.7\%)     & -- \\
            RRBP~\cite{zhou2019accelerate}          & & --             & 73.00\%             & -3.13\%           & 91.00\%           & --                 & -- (54.5\%)\\ 
            \textbf{AutoBot (Ours)}                & \checkmark  & \textbf{47.51\%} & \textbf{76.63\%}  & \textbf{+0.50\%} & \textbf{92.95\%}  & \textbf{1.97B (52.0\%)} &   \textbf{16.73M (34.5\%)} \\
            ITPruner~\cite{Info_flow21_PAMI} & \checkmark & 0.10\%\textsuperscript{*}     & 75.28\%             & -0.85\%   & --     & 1.94B (52.8\%)     & -- \\
            GDP-0.6~\cite{globalPruningIJCAI}  & \checkmark & --       & 71.19\%             & -4.94\%           & 90.71\%           & 1.88B (54.0\%)     & -- \\
            SCOP~\cite{SCOP}                        & & 4.26\%             & 75.26\%             & -0.87\%           & 92.53\%       & 1.85B (54.6\%)     & 12.29M (51.9\%) \\
            GAL-0.5-joint~\cite{GAL}                & & --             & 71.80\%             & -4.33\%           & 90.82\%           & 1.84B (55.0\%)     & 19.31M (24.5\%) \\
            ABCPruner~\cite{ABCPruner} & \checkmark & 0.10\%\textsuperscript{*}      & 73.52\%             & -2.61\%      & 91.51\%  & 1.79B (56.6\%)     & 11.24M (56.0\%) \\
            GAL-1~\cite{GAL}                        & & --             & 69.88\%             & -6.25\%           & 89.75\%           & 1.58B (61.3\%)     & 14.67M (42.6\%) \\
            LFPC~\cite{Cpruning_variousCriteria}    & & --             & 74.18\%             & -1.95\%           & 91.92\%           & 1.60B (61.4\%)     & -- \\
            GDP-0.5~\cite{globalPruningIJCAI} & \checkmark & --        & 69.58\%             & -6.55\%           & 90.14\%           & 1.57B (61.6\%)     & -- \\
            DCFF~\cite{Minbao2021DCFF}              & & --             & 75.60\%             & -0.53\%           & 92.55\%           & 1.52B (63.0\%)     & 11.05M (56.8\%) \\
            DCFF~\cite{Minbao2021DCFF}              & & --             & 74.85\%             & -1.28\%           & 92.41\%           & 1.38B (66.7\%)     & 11.81M (53.8\%) \\
            \textbf{AutoBot (Ours)}            & \checkmark & \textbf{14.71\%}   & \textbf{74.68\%}  & \textbf{-1.45\%}  & \textbf{92.20\%} & \textbf{1.14B (72.3\%)} &  \textbf{9.93M (61.2\%)} \\
            CURL~\cite{LuoW20}                & \checkmark & 0.10\%        & 73.39\%             & -2.74\%           & 91.46\%       & 1.13B (72.5\%)      & 6.67M (73.9\%)\\
            GAL-1-joint~\cite{GAL}                  & & --             & 69.31\%             & -6.82\%           & 89.12\%           & 1.11B (73.0\%)     & 10.21M (60.1\%) \\
            DCFF~\cite{Minbao2021DCFF}              & & --             & 73.81\%             & -2.32\%           & 91.59\%           & 1.02B (75.1\%)     & 6.56M (74.3\%) \\
            \specialrule{1pt}{1pt}{1pt} 
            \midrule[.5pt]
            \multicolumn{5}{l}{\textsuperscript{*}\footnotesize{this method train the pruned model from scratch, instead of finetuning}}
        \end{tabular}
    }
    \label{tab:imagenet_table}
\end{table*}
\subsection{Experimental Settings}
To demonstrate the efficiency of AutoBot on a variety of experimental setups, experiments are conducted on two popular benchmark datasets and five common CNN architectures, 1) CIFAR-10~\cite{cifar10} with VGG-16~\cite{vgg}, ResNet-56/110~\cite{resNet}, DenseNet~\cite{densenet}, and GoogLeNet~\cite{googlenet}, and 2) ILSVRC2012 (ImageNet)~\cite{deng2009imagenet} with ResNet-50. 

Experiments are performed within the \textit{PyTorch} and \textit{torchvision} frameworks~\cite{paszke2017automatic} under \textit{Intel(R) Xeon(R) Silver 4210R CPU 2.40GHz} and \textit{NVIDIA RTX 2080 Ti with 11GB} for GPU processing.

For CIFAR-10, we trained the bottlenecks for 200 iterations with a batch size of 64, a learning rate of 0.6 and $\beta$ equal to 5.5, and we finetuned the model for 200 epochs with the initial learning rate of 0.02 scheduled by cosine annealing scheduler and with a batch size of 256. 
For ImageNet, we trained the bottlenecks for 3000 iterations with a batch size of 64, a learning rate of 0.4 and $\beta$ equal to 13, and we finetuned the model for 200 epochs with a batch size of 512 and with the initial learning rate of 0.006 scheduled by cosine annealing scheduler. Bottlenecks are optimized via Adam optimizer.
All networks are retrained via the Stochastic Gradient Descent (SGD) optimizer, with momentum of 0.9 and decay factor of $2\times10^{-3}$ for CIFAR-10 and with momentum of 0.99 and decay factor of $1\times10^{-4}$ for ImageNet. 

\subsection{Evaluation Metrics}
We first evaluate the accuracy of the models. We measure it after finetuning, as is common in DNN pruning literature. However, in contrast to other works, we also measure it right after the pruning step (before finetuning) to show that our method effectively preserves the important filters compared to other methods. 
In addition, we adopt the FLOPs and number of parameters to measure the computational efficiency and model size.

\subsection{Automatic Pruning on CIFAR-10}
To demonstrate the improvement of our method, we firstly conduct automatic pruning with some of the most popular convolutional neural networks, namely VGG-16, ResNet-56/110, GoogLeNet, and DenseNet-40. Tab.~\ref{tab:cifar10_table} indicates experimental results with these architectures on CIFAR-10 for various number of FLOPs.\\
 
\noindent
{\bf VGG-16} We performed on VGG-16 architecture with three different pruning ratios. Tab.~\ref{tab:cifar10_table} demonstrates that AutoBot can efficiently preserve initial Top-1 accuracy before finetuning, even under the same FLOPs reduction (e.g. 82.73\% (proposed method) vs. 10.00\% from 65.4\% (HRank), 68.6\% (ITPruner), and 73.7\% (ABCPruner) of FLOPs reduction), thus leading to a SOTA accuracy after finetuning. For instance, we get 71.24\% and 93.62\% accuracy before and after finetuning respectively when reducing the FLOPs by 76.9\%. Our method even outperforms the baseline by 0.05\% and 0.23\% when reducing the FLOPs by 65.4\% and 53.7\%, respectively.
 As emphasized in Fig.~\ref{fig:fig2}, the per-layer filter pruning ratio is automatically determined by our method, according to the target FLOPs.\\
 
\begin{figure}[t]
\centering
\includegraphics[width=0.75\columnwidth]{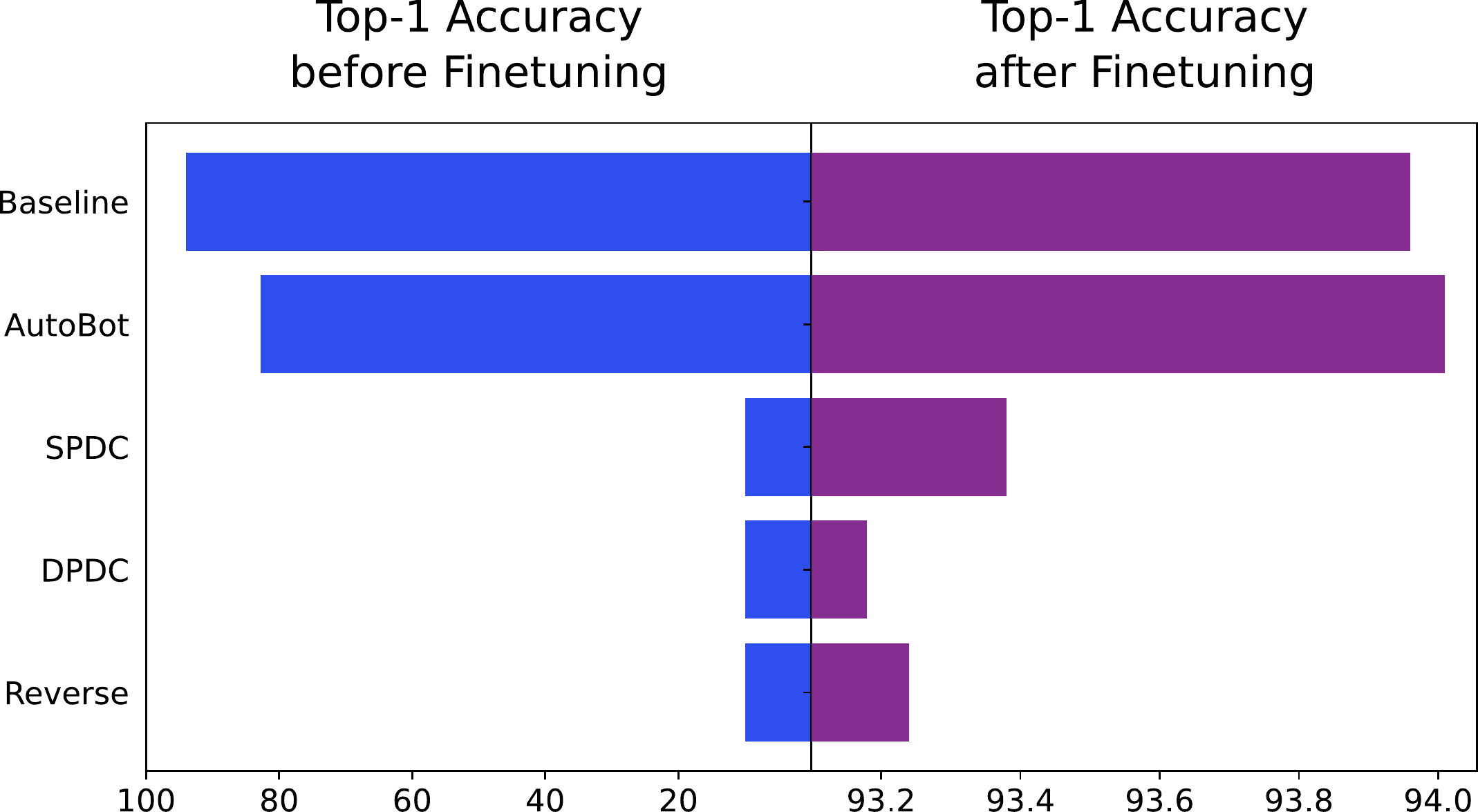}
\caption{Top-1 accuracy before and after finetuning for different pruning strategies, on VGG-16. The strategies are detailed in Sec.~\ref{subsec:initial_acc}}
\label{fig:fig4_init_acc}
\end{figure}

\begin{figure*}[t]
\centering
\includegraphics[width=0.9\textwidth]{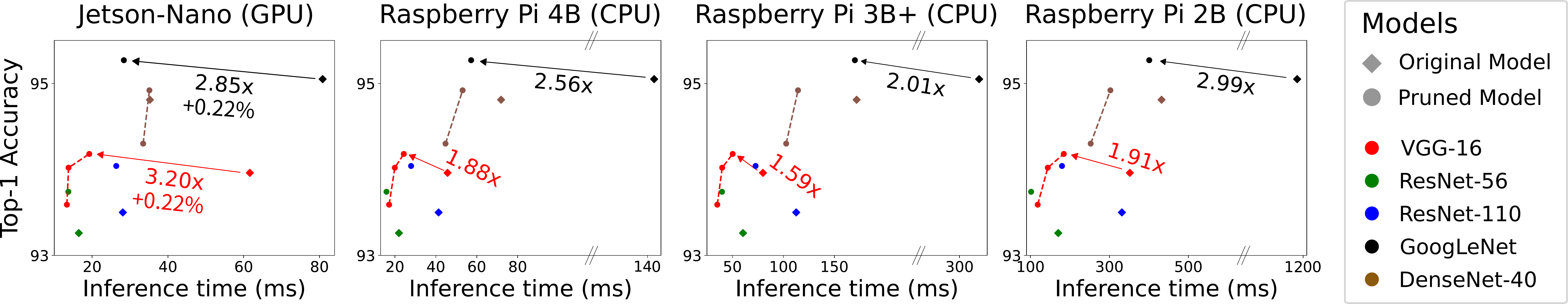}
\caption{Performance comparison between original and pruned models in terms of accuracy (x-axis) and inference time ($ms$) (y-axis) using five different networks on CIFAR-10. Top-left is better performance.}
\label{fig:fig3_deployment}
\end{figure*}
\noindent
{\bf ResNet} ResNet is an architecture characterized by its residual connections. Pruned model with our method can improve accuracy from 85.58\% before finetuning to 93.76\% after finetuning under a FLOPs reduction of 55.9\% for ResNet-56, and from 84.37\% before finetuning to 94.15\% after finetuning under a FLOPs reduction of 66.6\% for ResNet-110. Under similar or even smaller FLOPs, our approach accomplishes an excellent Top-1 accuracy compared to other existing magnitude-based or adaptive-based pruning methods and is beyond the baseline model's performance (93.27\% for ResNet-56 and 93.50\% for ResNet-110).\\

\noindent
{\bf GoogLeNet} GoogLeNet is a large architecture characterized by its parallel branches. Without any further processing, our initial accuracy of 90.18\% after pruning under a FLOPs reduction of 70.6\% (against 10\% for HRank and ABCPruner for the similar compression ratio) leads to the SOTA accuracy of 95.23\% after finetuning, outperforming recent papers such as DCFF and CC. Moreover, we also achieve a significant improvement in term of parameters reduction (73.1\%), although it is not the primary focus of our method.\\

\noindent
{\bf DenseNet-40} As ResNet, DenseNet-40 is an architecture based on residual connections. We experimented with two different target FLOPs, as shown in Tab.~\ref{tab:cifar10_table}. Notably, we got an accuracy of 83.2\% before finetuning and 94.41\% after finetuning under a FLOPs reduction of 55.4\%.

\subsection{Automatic Pruning on ImageNet}

To show the performance of our method on ILSVRC-2012, we chose the ResNet-50 architecture, made of 53 convolution layers followed by a fully-connected layer. Due to the complexity of this dataset (1,000 classes and millions of images), this task is more challenging than the compression of models on CIFAR-10. While existing pruning methods requiring to manually define the pruning ratio for each layer achieve reasonable performance, our global pruning method allows competitive results in all evaluation metrics including Top-1 and Top-5 accuracy, FLOPs reduction as well as number of parameters reduction, as reported in Tab.~\ref{tab:imagenet_table}. Under the high FLOPs compression of 72.3\%,  we obtain an accuracy of 74.68\%, outperforming recent works including GAL (69.31\%) and CURL (73.39\%) with a similar compression. And under the compression of 52\%, our method even outperforms the baseline by 0.5\% and leaves all the previous methods behind by at least 1\% by doing so. Therefore, the proposed method also works well on a complex dataset.

\subsection{Ablation Study}
\subsubsection{Impact of Preserving the Accuracy}
\label{subsec:initial_acc}

To highlight the impact of preserving the accuracy during the pruning process, we compare the accuracy before and after finetuning of AutoBot with different pruning strategies in Fig.~\ref{fig:fig4_init_acc}.
To show the superiority of an architecture found by preserving the accuracy compared to a manual design, a comparison study is conducted by manually designing three different strategies: 1) Same Pruning, Different Channels (SPDC), 2) Different Pruning, Different Channels (DPDC), and 3) Reverse.

DPDC has the same FLOPs as the architecture found by AutoBot but uses a different per-layer pruning ratio proposed by Lin \etal~\cite{hrank}.
To show the impact of a bad initial accuracy for finetuning, we propose the SPDC strategy that has the same per-layer pruning ratio as the architecture found by AutoBot but with randomly selected filters.
We also propose to reverse the order of importance of the filters selected by AutoBot such that only the less important filters are pruned. By doing so, we can better appreciate the importance of the scores returned by AutoBot. In Fig.~\ref{fig:fig4_init_acc}, we define this strategy as Reverse. This strategy gives a different per-layer pruning ratio than the architecture found by AutoBot.
We evaluate the three strategies on VGG-16 with a pruning ratio of 65.4\%, and we use the same finetuning conditions for all of them. We select the best accuracy among 3 runs. As shown in Fig.~\ref{fig:fig4_init_acc}, these three different strategies give an initial accuracy of 10\%. While the DPDC strategy gives an accuracy of 93.18\% after finetuning, the SPDC strategy displays 93.38\% accuracy, thus showing that an architecture found by preserving the initial accuracy
gives better performance. Meanwhile, the Reverse strategy obtains 93.24\%, which is surprisingly better than the hand-made architecture but, as expected, it underperforms the architecture found by AutoBot, even if we apply the SPDC strategy. 

\subsubsection{Deployment Test}
To highlight the improvement in real situations, we compare the inference speed-up of our compressed networks deployed on GPU-based (NVIDIA Jetson Nano) and CPU-based (Raspberry Pi 4, Raspberry Pi 3, and Raspberry Pi 2) edge devices. Specifications of these devices are available in Tab.~\ref{tab:specifictation} in the appendix. The pruned models are converted into ONNX format. Fig.~\ref{fig:fig3_deployment} shows the comparison study for inference times between the original pre-trained models and our compressed models. We can show that inference time for our pruned models is improved in every target edge device (e.g. GoogleNet is 2.85$\times$ faster on Jetson-Nano and 2.56$\times$ faster on Raspberry Pi 4B with 0.22\% increased accuracy). Especially, the speed is significantly better on GPU-based devices for single sequence of layers models (e.g. VGG-16 and GoogLeNet) whereas it improved the most on CPU-based devices for models with skip connections. More detailed results are available in Tab.~\ref{tab:deployment}.


\section{Limitations}
\label{sec:limitation}

While pruning with Autobot is a fast process, finding the hyper-parameters that most efficiently preserve the accuracy requires a hyper-parameter optimization step. However, our experiments highlight the relative stability of these hyper-parameters for different models on the same dataset. For instance, all our results on CIFAR10 presented in Tab.~\ref{tab:cifar10_table} were obtained with the same hyper-parameters.

For complex architectures, manually placing the bottlenecks can be challenging as it requires identifying which operations must be pruned together. It is interesting to notice that this could be solved with automation as these dependencies follow simple rules (e.g., in case of a skip connection, if the branches are summed then they should be pruned together).


\section{Conclusion}
\label{sec:conclusion}
In this paper, we introduced AutoBot, a novel automatic pruning method focusing on FLOPs reduction. To determine which filters to prune, AutoBot employs trainable bottlenecks designed to preserve the channels that maximize the model accuracy while minimizing the FLOPs. Notably, these bottlenecks only require one epoch on 25.6\% (CIFAR-10) or 15.0\% (ILSVRC2012) of the dataset to be trained. Extensive experiments on various CNN architectures demonstrate that the proposed method is superior to previous channel pruning methods both before and after finetuning. Our paper is the first to compare accuracy before finetuning. 


\bibliography{aaai23}

\clearpage

\appendix

\section{Reduced training data on ILSVRC2012}
\label{sec:reduced_data_imagenet}
\begin{figure}[h]
    \centering
    \includegraphics[width=0.7\columnwidth]{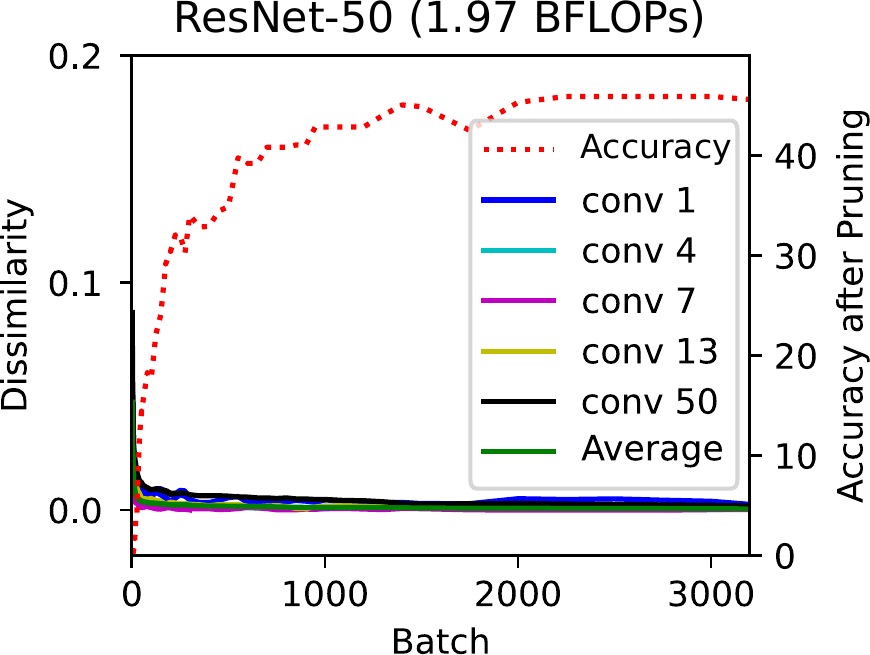}
    \caption{Evolution of accuracy after pruning (before finetuning) and of dissimilarity between filters ranking (normalised Kendall tau distance) when increasing the number of batches, on ILSVRC2012.}
    \label{fig:acc_after_pruning_and_dissimilarity_imagenet}
\end{figure}

As discussed in the Pruning Strategy sub-section, AutoBot only requires a small subset of the dataset to converge optimally. This section demonstrates the same phenomena with a supplementary experiment on ILSVRC2012. Fig.~\ref{fig:acc_after_pruning_and_dissimilarity_imagenet} shows the accuracy after pruning (before finetuning) and the dissimilarity between filters ranking between two parameters updates when pruning ResNet-50 on ILSVRC2012. Dissimilarity is computed using the normalised Kendall tau distance, which is a common tool to measure dissimilarity between rankings.

In this experiment, we can observe that the accuracy after pruning can converge at around 3000 batches (15.0\% of the training dataset), while the dissimilarity between filters is also stable throughout the network.

\section{Pruning process time}
\label{sec:pruning_process_time}

Tab.~\ref{tab:pruning_process_time} shows the time it takes for the whole pruning process with AutoBot, including the model and data loading, the bottlenecks training, the computation of the optimal threshold, and the actual pruning of the filters. On CIFAR-10, regardless of the architectures' heaviness, AutoBot can provide the optimal pruning ratio layer-by-layer under the target FLOPs within a minute. Even on ILSVRC2012, which requires larger computational times in general, AutoBot can efficiently prune ResNet-50 in around 10 minutes.

To summarize, this table highlights that the model compression is relatively fast thanks to the fast convergence of the bottlenecks.

\begin{table}[h!]
    \begin{minipage}[b]{80mm}
    \caption{Pruning process time on NVIDIA RTX 2080 Ti, in GPU hours}
    \centering
    \scalebox{0.6}{
        \begin{tabular}{c||c|c|c|c|c|c}
            \specialrule{1pt}{1pt}{1pt}
            \hline
            Dataset      &   \multicolumn{5}{c|}{CIFAR-10} & ILSVRC2012 \\ \hline
            Model        &  VGG-16  & ResNet-56 &  ResNet-110 & GoogleNet & DenseNet-40 & ResNet-50  \\ \hline
            GPU hours    &   0.005  &   0.009   &     0.011   &   0.011   &    0.013    &    0.182   \\ \hline
            \specialrule{1pt}{1pt}{1pt}
        \end{tabular}
    }
    \label{tab:pruning_process_time}
    \end{minipage}
\end{table}

\section{Hardware specification}
\label{sec:hardware}

Tab.~\ref{tab:specifictation} summarize the hardware specifications used for the inference speed tests in our paper.

\begin{table}[!h]
    \begin{minipage}[b]{80mm}
    \caption{The specification of the hardware platforms deployed in our paper.}
    \centering
    \scalebox{0.75}{
        \begin{tabular}{c|ccc}
            \specialrule{1pt}{1pt}{1pt}
            \hline 
            Platform         &       CPU                                     &   GPU     & Memory    \\ \hline
            Jetson-Nano      & \makecell{Quad core Cortex \\-A57 @ 1.43GHz} &   \makecell{128-core \\ Maxwells$\textsuperscript{TM}$ $\mu$A} &  4GB LPDDR4                  \\
            Raspberry Pi 4B  & \makecell{Quad core Cortex \\-A72 @ 1.5GHz}   &  No GPGPU &  4GB LPDDR4 \\
            Raspberry Pi 3B+ & \makecell{Quad core Cortex \\-A53 @ 1.4GHz}   &  No GPGPU &  1GB LPDDR2 \\
            Raspberry Pi 2B  & \makecell{Quad core Cortex \\-A7 @ 900 MHz}  &  No GPGPU &  1GB SDRAM  \\
            \specialrule{1pt}{1pt}{1pt}
        \end{tabular}
    }
    \label{tab:specifictation}
    \end{minipage}
\end{table}

\section{Inference speed detailed results}
\label{sec:inference_speed}

Tab.~\ref{tab:deployment} show the detailed results of the inference speed tests on CIFAR10 conducted in this work.

\begin{table*}[!h]
    \centering
    \caption{Deployment test on different hardware platforms with our pruned model. Inference time for original model ($ms$) ~$\rightarrow$~ inference time for pruned model ($ms$).}
    \label{tab:deployment}
    \scalebox{0.73}{
        \begin{tabular}{cc|cccc}
            \specialrule{1pt}{1pt}{1pt}
            \hline
            &   & \multicolumn{4}{c}{Hardware (Processor)}  \\ \hline
            Model         & FLOPs      &       Jetson-Nano (GPU)     &       Raspberry Pi 4B (CPU)     & Raspberry Pi 3B+ (CPU) &       Raspberry Pi 2B (CPU)      \\ \hline
            VGG-16     &   73.71M   &    61.63 $\rightarrow$ 13.33 (\textbf{$\times$~4.62})     &       45.73 $\rightarrow$ 17.16 (\textbf{$\times$~2.66})      
            &       79.98 $\rightarrow$ 35.17 (\textbf{$\times$~2.27}) &    351.77 $\rightarrow$ 118.36 (\textbf{$\times$~2.97})  \\
            VGG-16     &   108.61M  &    61.63 $\rightarrow$ 13.77 (\textbf{$\times$~4.48})     &       45.73 $\rightarrow$ 19.95 (\textbf{$\times$~2.29})     
            &       79.98 $\rightarrow$ 39.99 (\textbf{$\times$~2.00}) &    351.77 $\rightarrow$ 143.95 (\textbf{$\times$~2.44})  \\
            VGG-16     &   145.55M  &    61.63 $\rightarrow$ 19.24 (\textbf{$\times$~3.20})     &       45.73 $\rightarrow$ 24.33 (\textbf{$\times$~1.88})      
            &       79.98 $\rightarrow$ 50.27 (\textbf{$\times$~1.59}) &    351.77 $\rightarrow$ 184.47 (\textbf{$\times$~1.91})  \\
            ResNet-56     & 55.94M     &   16.47  $\rightarrow$ 13.71 (\textbf{$\times$~1.20})     &       21.95 $\rightarrow$ 15.88 (\textbf{$\times$~1.38})      
            &       60.42 $\rightarrow$ 39.78 (\textbf{$\times$~1.52}) &    170.46 $\rightarrow$ 101.70 (\textbf{$\times$~1.68})  \\
            ResNet-110    & 85.30M     &    28.10 $\rightarrow$ 26.36 (\textbf{$\times$~1.07})     &       41.35 $\rightarrow$ 27.90 (\textbf{$\times$~1.48})      
            &      112.57 $\rightarrow$ 72.71 (\textbf{$\times$~1.55}) &    331.60 $\rightarrow$ 179.91 (\textbf{$\times$~1.84})  \\
            GoogLeNet     & 0.45B      &    80.84 $\rightarrow$ 28.37 (\textbf{$\times$~2.85})     &      146.68 $\rightarrow$ 57.25 (\textbf{$\times$~2.56})      
            &      342.23 $\rightarrow$ 170.17 (\textbf{$\times$~2.01}) &   1,197.65 $\rightarrow$ 400.89 (\textbf{$\times$~2.99})  \\
            DenseNet-40   & 129.13M     &    35.25 $\rightarrow$ 33.46 (\textbf{$\times$~1.05})     &      71.87 $\rightarrow$ 44.73 (\textbf{$\times$~1.61})      
            &      171.86 $\rightarrow$ 102.75 (\textbf{$\times$~1.67}) &     432.03 $\rightarrow$ 252.63 (\textbf{$\times$~1.71})  \\ 
            DenseNet-40   & 168.26M     &    35.25 $\rightarrow$ 35.11 (\textbf{$\times$~1.00})     &      71.87 $\rightarrow$ 53.08 (\textbf{$\times$~1.35})      
            &      171.86 $\rightarrow$ 114.37 (\textbf{$\times$~1.50}) &     432.03 $\rightarrow$ 302.49 (\textbf{$\times$~1.43})  \\ 
            \hline
        \end{tabular}
    }
\end{table*}

\end{document}